\documentclass{article} 
\usepackage{iclr2026_conference,times}
\iclrfinalcopy
\usepackage{enumitem}
\usepackage{amsmath}

\usepackage{graphicx}
\usepackage{booktabs}
\usepackage{multirow}
\usepackage{pifont}
\usepackage{caption,subcaption}
\usepackage{adjustbox}
\usepackage{soul, color, xcolor}
\definecolor{myorange}{RGB}{255,165,0}

\usepackage[table]{xcolor} 

\usepackage{tcolorbox}
\tcbuselibrary{skins, breakable}
\usepackage{xcolor}
\definecolor{mygray}{RGB}{192,192,192}

\usepackage{float}
\usepackage{xspace}
\tcbset{
  aibox/.style={
    width=\textwidth,
    top=10pt,
    colback=white,
    colframe=black,
    colbacktitle=black,
    enhanced,
    center,
    attach boxed title to top center={yshift=-0.1in},
    boxed title style={boxrule=0pt,colframe=white,},
  }
}
\newtcolorbox{AIbox}[2][]{aibox,title=#2,#1}
\usepackage{listings}
\definecolor{myblue}{RGB}{100, 150, 200}
\definecolor{mygreen}{RGB}{80, 160, 80}

\definecolor{darkgreen}{rgb}{0.0, 0.5, 0.0}
\definecolor{darkgray}{gray}{0.4}
\definecolor{maroon}{rgb}{0.5, 0.0, 0.0}
\definecolor{navy}{rgb}{0.0, 0.0, 0.5}
\definecolor{teal}{rgb}{0.0, 0.5, 0.5}

\usepackage{amsmath,amsfonts,bm}

\def\eqref#1{equation~\ref{#1}}

\def\1{\bm{1}}

\def\ve{{\bm{e}}}

\def\vx{{\bm{x}}}

\DeclareMathAlphabet{\mathsfit}{\encodingdefault}{\sfdefault}{m}{sl}
\SetMathAlphabet{\mathsfit}{bold}{\encodingdefault}{\sfdefault}{bx}{n}

\usepackage{hyperref}
\usepackage{url}
\usepackage{algorithm}
\usepackage{algorithmic} 

\title{From Long News to Accurate Forecast: Importance-Aware Fusion and PRM-Guided Reflection for Time Series Forecasting}

\author{
Mingyang LIU$^{1*}$, Qingcan Kang$^{2\dagger}$,\textbf{Yuke WANG}$^{1}$, Shixiong Kai$^{2}$, Kaichao Liang$^{2}$, 
\\\textbf{Hui-Ling Zhen}$^{2}$, \textbf{Tao Zhong}$^{2}$,
~\textbf{Mingxuan Yuan}$^{2}$, \textbf{Linqi Song}$^{1\dagger}$\\
     $^{1}$Department of Computer Science, City University of Hong Kong \\
     $^{2}$Huawei Noah's Ark Lab \\
     \texttt{mingyaliu8-c@my.cityu.edu.hk}\\
     \texttt{kangqingcan@huawei.com}\\
    \texttt{linqi.song@cityu.edu.hk}
}

\begin{document}

\maketitle

\def\customfootnotetext#1#2{{%
		\let\thefootnote\relax
		\footnotetext[#1]{#2}}}

\customfootnotetext{1}{\textsuperscript{*} Work done as an intern in Huawei Noah's Ark Lab.}
\customfootnotetext{2}{{$\dagger$} Corresponding authors.}

\begin{abstract}
Incorporating news into time series forecasting is appealing because news can reveal abrupt exogenous events that historical values alone cannot recover. However, existing LLM-based news-forecasting pipelines face two practical limitations: relevant news articles often exceed the model's context window, and iterative retrieval of supplementary news is typically unguided, leading to redundant updates and slow convergence. We address these issues with a novel framework that combines importance-aware news compression and process-level retrieval supervision. First, we train an importance reward model that estimates the forecasting utility of each article and uses this signal to allocate compression budgets during sequential pairwise fusion, preserving informative content within a fixed context limit. Second, we introduce a process reward model (PRM) that ranks multiple supplementary-news candidates conditioned on the current error profile and the history of previously selected articles, replacing one-shot blind retrieval with quality-controlled selection. Both components are trained offline using historical data with ground truth; inference uses the frozen filtering logic and compression modules without any reflection loop. Experiments on finance, energy, traffic, and bitcoin forecasting benchmarks show that our method improves prediction accuracy over strong baselines, significantly reduces the number of refinement iterations compared to the iterative baseline, and remains effective when relevant articles span thousands of tokens.
\end{abstract}
\section{Introduction}
\vspace{-2mm}

Time series forecasting supports decision making in domains such as finance, energy, and transportation \citep{alghamdi2019forecasting, fildes2022retail, gross1987short}. Although modern forecasting models can capture rich temporal patterns from historical observations, they remain vulnerable to abrupt distribution shifts caused by exogenous events, such as policy changes, natural disasters, or market-moving announcements. These events are often weakly reflected, or not reflected at all, in the numerical history available at prediction time. This limitation motivates forecasting systems that can incorporate external textual evidence, especially news reports, into the prediction process.

News provides timely descriptions of latent drivers behind future time-series movements. Recent LLM-based forecasting pipelines have started to exploit this signal by prompting a reasoning agent to filter relevant articles and a separate evaluation agent to analyze prediction errors and propose additional retrieval queries \citep{wang2024news}. This iterative design is attractive because it converts missing-context errors into language feedback that can be used to refine the news set. However, two obstacles limit its effectiveness in realistic settings.

First, the retrieved news may be long. Articles such as policy briefings, earnings-call transcripts, and incident reports can easily exceed the context budget of the forecasting model. Naive truncation is unreliable because it discards potentially causal details, while generic summarization or prompt-compression methods \citep{jiang2023llmlingua, jiang2024longllmlingua} are not optimized for preserving information that is specifically useful for forecasting. As a result, there is a mismatch between context efficiency and task relevance. Second, the refinement loop is weakly supervised. In prior iterative pipelines, a reflection may indicate a missing category of news, but the subsequent supplementary retrieval is effectively accepted without a dedicated quality-control mechanism. This makes the update trajectory sensitive to noisy or redundant articles and can lead to many expensive refinement rounds before the filtering logic stabilizes.

We address these two issues with a novel framework that separates \emph{offline refinement} from \emph{online deployment}. During offline training, where ground-truth targets are available, we improve the news-processing pipeline in two ways. We first introduce an importance-aware fusion module that learns an article-level reward model and allocates compression budgets according to the estimated utility of each article for forecasting. We then introduce a process reward model (PRM) that ranks multiple supplementary-news candidates conditioned on the current forecasting errors and the history of previous selections, so that the refinement loop can prefer candidates that are more likely to reduce future error. At inference time, the refined filtering logic and the trained compression components are frozen, and prediction proceeds without iterative reflection. Experiments on finance, energy, traffic, and bitcoin forecasting benchmarks demonstrate that our method improves prediction accuracy over strong baselines, reduces the number of refinement iterations by up to 37.6\% with an average reduction of 24.8\%, and remains effective when relevant articles span thousands of tokens.

\paragraph{Contributions.}
Our main contributions are summarized as follows:
\begin{itemize}[itemsep=0mm, leftmargin=6mm, topsep=0mm]
    \item We formulate news-augmented forecasting as an offline-refined pipeline that explicitly addresses two bottlenecks of prior LLM-based approaches: long-document compression and uncontrolled supplementary retrieval.
    \item We propose an importance-aware fusion module that learns forecasting-oriented article utilities and uses them to adapt compression rates during sequential pairwise fusion, enabling long news to be incorporated under a limited context budget.
    \item We introduce a PRM-guided refinement strategy for supplementary news selection, replacing one-shot retrieval with candidate ranking based on process-level supervision from error-driven trajectories.
\end{itemize}

\section{Related Work}
\label{sec:related}
\vspace{-2mm}

\textbf{Time series forecasting with external text.}\quad
Classical forecasting models rely primarily on historical numerical observations \citep{chen2004load, dudek2015short, huang2003short, kalekar2004time, papalexopoulos1990regression}. Deep neural architectures improved the modeling of long-range and nonlinear dependencies \citep{li2019enhancing, liu2017short, nie2022time, torres2021deep, wu2021autoformer, xia2021stacked, zhou2021informer, zhou2022fedformer}, and large-scale pre-training further strengthened transfer to new tasks \citep{cao2023tempo, jin2023large, wu2022timesnet, yeh2023toward}. A parallel line of research augments forecasting with textual evidence such as news, reports, or social media, especially in finance and energy \citep{cecchini2010making, schumaker2009textual, schumaker2010discrete, bai2024news, obst2021adaptive}. Early methods rely on manually engineered sentiment features or event indicators \citep{obst2021adaptive, chowdhary2020natural}, which limits their ability to capture long-range semantics and cross-document interactions.

\textbf{LLM agents and iterative refinement.}\quad
Recent work adapts LLMs to time series by tokenizing numerical sequences or reprogramming prompts \citep{gruver2023large, cao2023tempo, jin2023time, rasul2023lag}. The closest prior work is \citep{wang2024news}, which uses LLM agents to filter news and iteratively refines the filtering strategy based on forecast errors. That framework demonstrates the value of agentic feedback loops, but it assumes news documents are short and that each supplementary retrieval is unconditionally accepted. Our work builds on this agentic formulation and introduces two complementary mechanisms: an importance-aware compression module that respects context budgets, and a quality-controlled selection mechanism for supplementary news.

\textbf{Compression and process supervision.}\quad
Prompt compression methods such as LLMLingua \citep{jiang2023llmlingua}, LongLLMLingua \citep{jiang2024longllmlingua}, and RECOMP \citep{xu2024recomp} reduce input length for general-purpose LLM tasks, but the notion of importance is not tied to forecasting error. Process reward models (PRMs) \citep{lightman2023verify, zeng2025versaprm, zhang2025bidirectional, zheng2025prmsurvey} provide step-wise supervision for reasoning tasks, yet they are typically used to score rationales, not to rank retrieval candidates. Our work connects these two lines: we train a forecasting‑aligned reward model to allocate compression budgets, and a PRM to score candidate news articles based on their expected impact on prediction error. This joint design addresses a previously underexplored bottleneck in news‑augmented forecasting.
\begin{figure*}[t]
    \centering
    \includegraphics[width=1\linewidth]{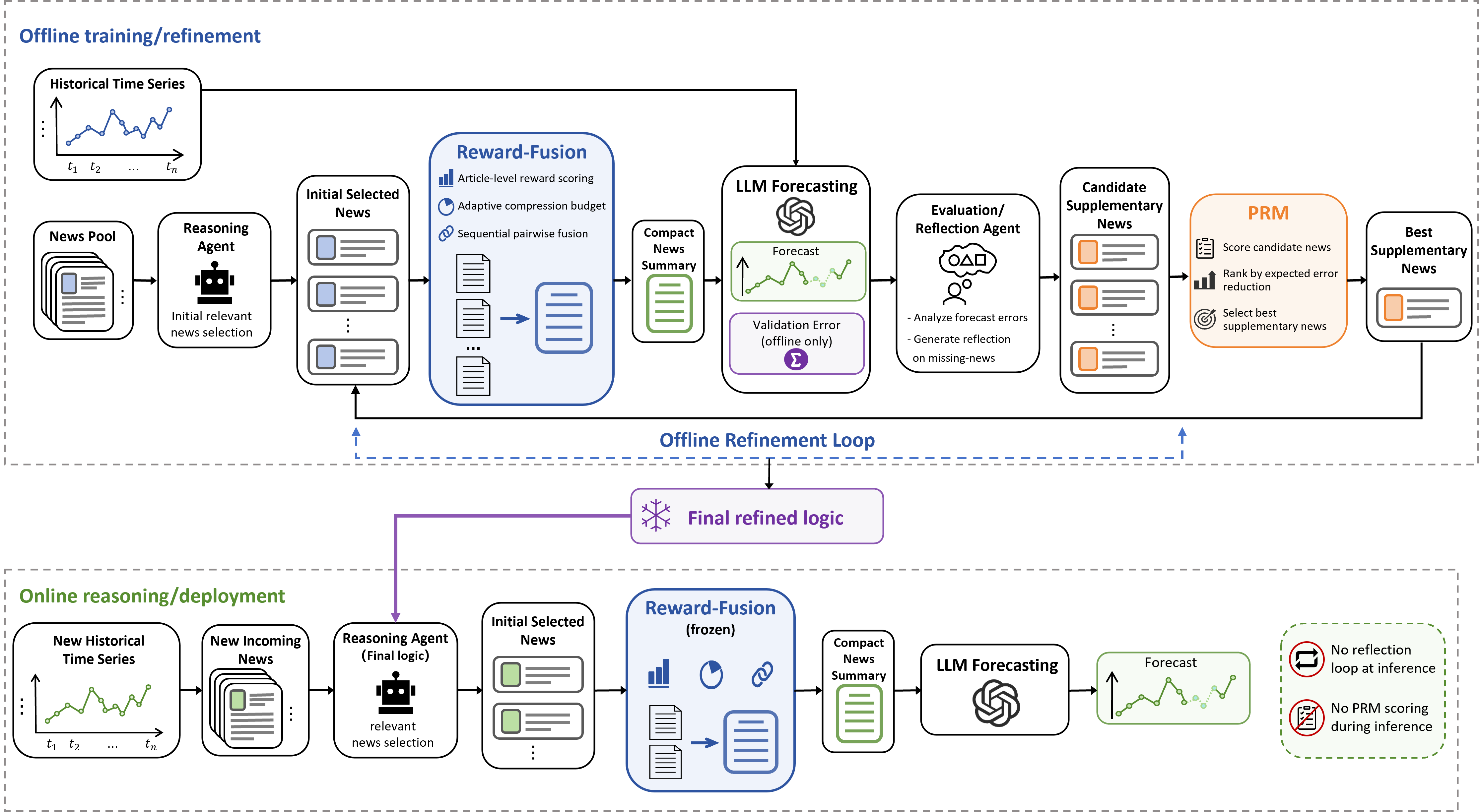}
        \caption{Overall Framework of Offline Refinement and Online Deployment. The framework consists of two phases: offline training/refinement and online reasoning/deployment. In the offline phase, historical time series and candidate news are first processed by the Reward-Fusion module, which performs article-level reward scoring, adaptive compression budgeting, and sequential pairwise fusion to produce a compact summary for forecasting. Based on validation errors, a reflection signal is generated to identify missing information, and a PRM module scores supplementary candidate news by their expected error reduction, selecting the best article to refine the filtering logic. This refinement loop is repeated until convergence, yielding the final refined logic. In the online phase, the refined logic and frozen Reward-Fusion module are directly applied to retrieve, compress, and fuse relevant news for forecasting, without reflection loops or PRM scoring during inference.}
    \label{fig:overall}
    \vspace{-20pt}
\end{figure*}

\section{Methodology}
\label{sec:method}

We consider forecasting with historical observations and a large pool of time-aligned news articles. The main difficulty is not only to identify relevant articles, but also to integrate them under limited context while avoiding noisy iterative updates. Our framework therefore contains four components: (i) an LLM forecasting backbone, (ii) a reasoning agent for initial news filtering and error-driven refinement, (iii) an importance-aware fusion module for compressing long news into a compact conditioning context, and (iv) a PRM that ranks supplementary-news candidates during refinement. The key design principle is to perform expensive search and supervision \emph{offline}, where ground-truth targets are available, and to deploy only the frozen filtering and compression logic \emph{online}. \figurename~\ref{fig:overall} summarizes the pipeline.

\subsection{Problem Formulation}
\label{sec:problem_formulation}

Let $\mathbf{x}_{1:T} = (x_1, \ldots, x_T)$ denote a sequence of univariate or multivariate time series observations up to the current time $T$. The task is to forecast the next $H$ values $\mathbf{x}_{T+1:T+H}$. In addition to the numerical history, we have access to a collection of news articles $\mathcal{U} = \{u_1, \ldots, u_L\}$ whose publication timestamps fall within a retrieval window before the forecast time. The articles can be of arbitrary length, often exceeding the context limit of the forecasting model.

A reasoning agent first selects a subset $\mathcal{N}^{(0)} \subseteq \mathcal{U}$ of potentially relevant articles, which is then compressed into a textual summary $S^{(0)}$. The forecasting model $F_\psi$ conditioned on both the numerical history and the summary produces the prediction:
\[
\hat{\mathbf{x}}_{T+1:T+H} = F_\psi(\mathbf{x}_{1:T}, S^{(0)}).
\]

During offline refinement, we have access to ground-truth targets for a historical validation set. The prediction error is used to generate a textual reflection $r_t$ that describes missing evidence. This reflection triggers a retrieval of supplementary candidates, from which a process reward model (PRM) selects a news article to update the current news set. The refinement loop repeats until the validation improvement becomes negligible or a fixed budget is exhausted. The final refined filtering logic and trained compression modules are then deployed for online inference without any reflection loop.

\subsection{Preliminary: Time Series Forecasting with LLMs}
\label{sec:fintune_llm}
\vspace{-2mm}

Following \citep{wang2024news}, we cast forecasting as conditional sequence generation. A pre-trained LLM receives a prompt containing the historical series, task metadata, and a compressed textual summary of relevant news, and outputs the next $H$ values. If the numerical history is tokenized as $\vx_{1:T}$ and the news summary is tokenized as $\ve_{1:U}$, the model learns $P_\psi(\vx_{T+1:T+H} \mid \vx_{1:T}, \ve_{1:U})$.

We fine-tune the backbone with supervised instruction tuning and LoRA \citep{hu2022lora}. The role of our method is not to change the forecasting backbone itself, but to improve the quality and efficiency of the textual context $\ve_{1:U}$ that conditions the forecast.

\subsection{Iterative News Filtering with PRM-Guided Supplementary News Selection}
\label{sec:reasoning_agent}
\vspace{-2mm}

We adopt the initial reasoning stage of \citep{wang2024news} to obtain a seed news set $\mathcal{N}^{(0)} = \{N_1,\ldots,N_{K_0}\}$. The agent uses task descriptions, few-shot examples, and chain-of-thought prompting \citep{brown2020language, wei2022chain} to identify news categories likely to influence the target series and returns structured filtering outputs.

The main modification concerns the refinement stage. Let $\mathcal{N}^{(t)}$ be the news set after iteration $t$, and let $r_t$ denote the reflection produced from the current validation errors. Instead of accepting a single supplementary article directly from the reasoning agent, we retrieve a candidate pool $\mathcal{C}_t = \{c_t^{(1)}, \ldots, c_t^{(M)}\}$ that matches the reflection. The PRM then assigns a utility score to each candidate, and the highest-scoring article is added to the news set:
\[
n_t^* = \arg\max_{c \in \mathcal{C}_t} P_\phi(\text{beneficial} \mid h_t, \tau_{<t}, c), \qquad
\mathcal{N}^{(t+1)} = \mathcal{N}^{(t)} \cup \{n_t^*\},
\]
where $h_t$ summarizes the current task context and $\tau_{<t}$ is the history of previously selected supplementary articles. This design turns iterative retrieval into a ranking problem with explicit supervision, rather than an uncontrolled one-shot update.

The refinement procedure is executed only offline. Once the filtering logic has been refined on historical data, we keep the resulting logic fixed for deployment and do not invoke reflection or PRM scoring at inference time.

\subsection{Process Reward Model for Supplementary News Selection}
\label{sec:prm}
\vspace{-2mm}

A central limitation of prior iterative retrieval is the absence of step-level supervision for supplementary-news selection. We address this by training a PRM that estimates whether adding a candidate article at the current refinement step is likely to improve forecasting performance. The PRM outputs a scalar score used for candidate ranking.

\paragraph{Trajectory definition.}
Let $\tau = (n_1, n_2, \dots, n_T)$ denote a trajectory of supplementary selections. After adding $n_t$ to the news set, we re-compress the current news context, update the forecasting model, and evaluate on the validation set. The step-wise gain is defined as $\Delta_t := \text{RMSE}_{t-1} - \text{RMSE}_t$, where $\Delta_t > 0$ indicates that the newly selected article improves validation RMSE. The PRM models the probability that a candidate is beneficial can be expressed as $P_\phi(\text{beneficial} \mid h_t, \tau_{<t}, c)$, where $h_t$ contains the current context, including the residual pattern of the forecasting model, metadata of the task, and the current news set.

\paragraph{Training data construction.}
We construct supervision from offline trajectories on a held-out split of the training data. For each reflection $r_t$, the reasoning agent retrieves $M$ candidates. Each candidate is inserted temporarily into the pipeline, and we measure the corresponding gain $\Delta_t^{(m)}$. The binary label is $y_t^{(m)} = \mathbf{1}[\Delta_t^{(m)} > 0]$.
The PRM input concatenates: (1) Task description (horizon, domain, background); (2) Recent prediction errors (e.g., time‑aligned residuals); (3) The history $\tau_{<t}$ and the associated $\Delta$ values; (4) The candidate article $n_t^{(m)}$ (title and summary).

We train a binary classifier with cross-entropy loss. In practice, the classifier can be implemented as a lightweight language model or as a linear head on top of a frozen text encoder. The predicted sigmoid probability is used directly as the candidate score.

\paragraph{PRM-guided selection and logic update.}
During offline refinement, the reasoning agent retrieves a candidate set for each reflection \(r_t\), and the PRM ranks these candidates using the current context \(h_t\) and history \(\tau_{<t}\). The selected article \(n_t^*\) is merged into the news set, after which the forecasting pipeline is re-evaluated. If the update improves validation performance, the selected article is also incorporated into the persistent filtering logic used by the reasoning agent. In this way, the PRM acts as a gate that reduces error-amplifying updates while preserving useful evidence for future retrieval. Algorithm~\ref{alg:prm_selection} summarizes the procedure.

\begin{algorithm}[htbp]
\caption{PRM-guided supplementary news selection and logic update}
\label{alg:prm_selection}
\renewcommand{\algorithmiccomment}[1]{\hfill $\triangleright$ #1}
\begin{algorithmic}[1]
\REQUIRE Current news set $\mathcal{N}^{(t)}$, forecasting model $\mathcal{M}^{(t)}$, reflection $r_t$, task context $h_t$, history $\tau_{<t}$, PRM $P_\phi$, reasoning agent $\text{RA}$, convergence threshold $\varepsilon$, max iterations $T_{\max}$
\STATE $t \leftarrow 0$
\STATE $\Delta_0 \leftarrow \infty$
\REPEAT
    \STATE $\mathcal{C} \leftarrow \text{RA}.\text{retrieve}(r_t)$ \COMMENT{$K$ candidate articles}
    \FOR{each candidate $c_j \in \mathcal{C}$}
        \STATE $s_j \leftarrow \text{sigmoid}\bigl(P_\phi(\text{beneficial} \mid h_t, \tau_{<t}, c_j)\bigr)$ \COMMENT{PRM scoring}
    \ENDFOR
    \STATE $n^*_t \leftarrow \arg\max_{c_j} s_j$ \COMMENT{select best candidate}
    \STATE $\mathcal{N}^{(t)}_{\text{new}} \leftarrow \{n^*_t\}$
    \STATE $\mathcal{N}^{(t+1)} \leftarrow \mathcal{N}^{(t)} \cup \mathcal{N}^{(t)}_{\text{new}}$
    \STATE $\mathcal{M}^{(t+1)} \leftarrow \text{train}\bigl(\mathcal{M}^{(t)}, \mathcal{N}^{(t+1)}\bigr)$ \COMMENT{re-compress and update forecast model}
    \STATE $\Delta_{t+1} \leftarrow \text{RMSE}(\mathcal{M}^{(t)}) - \text{RMSE}(\mathcal{M}^{(t+1)})$ \COMMENT{compute improvement}
    \IF{$\Delta_{t+1} > 0$}
        \STATE $\mathcal{L}^{(t+1)} \leftarrow \text{update\_logic}(\mathcal{L}^{(t)}, n^*_t)$ \COMMENT{refine persistent filtering logic}
    \ENDIF
    \STATE $t \leftarrow t+1$
\UNTIL{$\Delta_t < \varepsilon$ \textbf{or} $t \ge T_{\max}$}
\RETURN Refined logic $\mathcal{L}^{(t)}$, final model $\mathcal{M}^{(t)}$
\end{algorithmic}
\end{algorithm}
\vspace{-10pt}
\paragraph{Relation to the long‑term filtering logic.}
The selected article $n_t^*$ serves two roles: it may improve the current iteration, and it provides a supervised signal for refining the reasoning agent's persistent filtering logic. Over multiple iterations, this logic becomes better aligned with the forecasting objective. The PRM therefore reduces the chance that low-value supplementary articles corrupt the long-term retrieval behavior.

\subsection{Adaptive News Fusion via Importance-Aware Reward Model}
\label{sec:news_fusion}
\vspace{-2mm}

Relevant news articles can be substantially longer than the context available to the forecasting backbone. Generic compression methods reduce length, but they do not explicitly optimize for downstream forecasting utility. We therefore introduce an importance-aware fusion module that compresses long news while conditioning the compression rate on article-level utility estimates.

\paragraph{Sequential pairwise fusion.}
Given a filtered article list $\{N_1,\dots,N_K\}$, we compress the set into a summary $S$ through sequential pairwise fusion. At each step, the current summary is merged with the next article, and the target compression ratio of each input unit is determined by an estimated importance score. This design avoids forcing all articles through a uniform compression budget.

\paragraph{Importance-aware reward model.}
We train a reward model $R_\theta$ that maps an article $N$ to an importance score $\rho \in [0,1]$. To obtain supervision, we compare forecasting performance under different retention rates for the same article. Let $\text{RMSE}_{\text{high}}$ denote the validation RMSE when the article is aggressively compressed and $\text{RMSE}_{\text{low}}$ the RMSE when it is lightly compressed. We define an unnormalized utility signal as
\[
\tilde{\rho}(N) = \frac{\text{RMSE}_{\text{high}} - \text{RMSE}_{\text{low}}}{\text{RMSE}_{\text{no news}}},
\]
and normalize it to $[0,1]$ within each training split. A larger score indicates that preserving more information from the article is more beneficial for forecasting. We train $R_\theta$ with supervised regression on article representations extracted by a pre-trained model.

\paragraph{Fusion procedure.}
Algorithm~\ref{alg:fusion} gives the fusion procedure. Articles are ordered by timestamp or retrieval priority. For article $N_i$, the reward model outputs \(\rho_i = R_\theta(N_i)\), which is mapped to a retention ratio \(\alpha_i = f(\rho_i)\) through a monotone function \(f\). Higher utility implies larger retention. A controllable summarizer then merges the current summary with the next article under their respective retention budgets. The final summary $S$ is used as the textual context for forecasting. Compared with task-agnostic prompt compression, this procedure allocates the limited context budget according to estimated forecasting relevance.

\begin{algorithm}[htbp]
\caption{Sequential pairwise news fusion}
\label{alg:fusion}
\renewcommand{\algorithmiccomment}[1]{\hfill $\triangleright$ #1}
\begin{algorithmic}[1]
\REQUIRE List of articles $\mathcal{N} = [N_1, N_2, \dots, N_K]$, reward model $R_\theta$, summarizer $\text{SUMM}(\cdot, \cdot; \alpha_1, \alpha_2)$, mapping $f: [0,1] \to [0,1]$
\STATE $S \leftarrow N_1$ \COMMENT{initialize summary}
\STATE $\rho_{\text{sum}} \leftarrow R_\theta(N_1)$ \COMMENT{initial relevance score}
\FOR{$i = 2$ to $K$}
    \STATE $\rho_i \leftarrow R_\theta(N_i)$ \COMMENT{score new article}
    \STATE $\alpha_{\text{sum}} \leftarrow f(\rho_{\text{sum}})$ \COMMENT{map to retention weight}
    \STATE $\alpha_i \leftarrow f(\rho_i)$ \COMMENT{map to incorporation weight}
    \STATE $S \leftarrow \text{SUMM}(S, N_i; \alpha_{\text{sum}}, \alpha_i)$ \COMMENT{weighted fusion}
    \STATE $\rho_{\text{sum}} \leftarrow \text{aggregate}(\rho_{\text{sum}}, \rho_i)$ \COMMENT{e.g., maximum or weighted average}
\ENDFOR
\RETURN $S$
\end{algorithmic}
\end{algorithm}


\vspace{-10pt}
\paragraph{Training and deployment.}
The reward model is trained offline and then reused throughout refinement and deployment. During online inference, the same deterministic scoring-and-fusion procedure is applied, without additional supervision or search.

\subsection{Overall Pipeline}
\label{sec:pipeline}
\vspace{-2mm}

The overall framework has two phases: offline refinement and online inference, as illustrated in \figurename~\ref{fig:overall}. Offline refinement uses historical data with ground truth to improve the filtering logic and train the auxiliary scoring modules. The procedure is: (1) \textbf{Initial filtering}: the reasoning agent applies a default logic to select a candidate news set $\mathcal{N}^{(0)}$.
(2) \textbf{Adaptive compression}: the importance-aware reward model (Sec.~\ref{sec:news_fusion}) compresses $\mathcal{N}^{(0)}$ into a compact summary $S$.
(3) \textbf{Forecasting fine-tuning}: the LLM is fine-tuned via LoRA on the paired time series and summary $S$.
(4) \textbf{Evaluation and reflection}: the evaluation agent analyzes validation-set errors and generates a textual reflection $r_t$ indicating a missing news category.
(5) \textbf{PRM-guided supplementary selection}: the reasoning agent retrieves $K$ candidate articles matching $r_t$; the PRM scores them and selects the highest-scoring article $n_t^*$, which is merged into the news set.
(6) \textbf{Iteration}: steps (2)–(5) are repeated until the RMSE improvement $\Delta_t$ falls below a threshold or a maximum number of rounds is reached.

After convergence, we freeze the refined filtering logic, the trained reward models, and the forecasting backbone. Online inference contains only a single forward pipeline: retrieve news with the frozen logic, compress the selected news with the importance-aware fusion module, and generate the forecast. Since ground truth is unavailable at test time, no reflection loop or PRM scoring is used. In our implementation, Deepseek V3.2 is used for the reasoning, summarization, and reflection modules, while a smaller Qwen model is used for the PRM and the importance reward model.
\section{Experiments}
\begin{table*}[t]
\caption{RMSE comparison across different domains and models on true news dataset. 
\textbf{Bold} indicates the best result and \underline{underline} indicates the second best result.}
\label{tab:results1}
\centering
\renewcommand{\arraystretch}{1.8}
\resizebox{\textwidth}{!}{%
\begin{tabular}{@{}lccccccccc@{}}
\toprule
Domain & Nbeats & PatchTST & DLinear & LightGBM & Chronos & iTransformer & LoRA & TimeLLM & Our method \\
\midrule
Electricity & \underline{374.55} & 397.11 & 468.01 & 383.45 & 765.61 & 428.85 & 695.73 & 505.66 & \textbf{372.32}\\
Bitcoin & 1921.42 & 1495.61 & 2291.27 & \underline{1156.65} & 2714.50 & 2309.84 & 2378.36 & 2467.26 & \textbf{886.58} \\
Traffic & 39.04 & 38.36 & 51.83 & \underline{34.87} & 47.47 & 46.31 & 50.06 & 55.17 & \textbf{34.66} \\
Exchange & 7.12 & 7.16 & 7.47 & \underline{7.09} & 7.14 & 8.56 & 7.89 & 7.23 & \textbf{5.86} \\
\bottomrule
\end{tabular}
}
\vspace{-15pt}
\end{table*}
\label{sec:exp}

\subsection{Data preparation}
We evaluate on datasets where exogenous textual signals offer meaningful predictive value—i.e., series influenced by human activities and socially salient events rather than purely physical processes. This excludes physics-driven benchmarks such as ETT \citep{haoyietal-informer-2021} and Weather \citep{reynolds2007daily,hersbach2020era5}, which are less suitable for studying event-conditioned auxiliary signals. Following this principle, we consider four domains: electricity demand \citep{godahewa2021monash}, Bitcoin price \citep{godahewa2021monash}, traffic flow \citep{maggie2017webtraffic}, and foreign exchange rates \citep{lai2018modeling}. The electricity, traffic, and foreign exchange datasets are adopted from \citep{wang2024news}; the Bitcoin price series is collected from investing.com \citep{investing_bitcoin_data}. Together, these benchmarks span 30-minute, hourly, and daily frequencies, enabling evaluation across varying temporal granularities.

For textual side information, we use real-world news corpora associated with the four domains provided by \citep{wang2024news}. To ensure strict separation between training and evaluation data while maintaining sufficient training scale, we also construct a large synthetic news corpus using DeepSeek-V3.2 \citep{deepseekai2025deepseekv32pushingfrontieropen} conditioned on real temporal information. The synthetic news is used exclusively for training the reward model and the process reward model (PRM), while evaluation is conducted only on real news. This protocol reduces the risk of train--test contamination, preserves realistic testing conditions, and provides a controlled and scalable source of textual supervision for model training. Table \ref{tab:news_length} provides detailed information on the length of news items in the test set. Please refer to the appendix for further details.

\subsection{Baseline}
To evaluate the performance of the proposed optimization approach, we carry out comprehensive experiments across a broad range of representative time-series forecasting models. In particular, the selected base forecasters fall into two groups. The first group consists of history-based forecasting models, including N-BEATS \citep{oreshkin2019n}, PatchTST \citep{nie2022time}, DLinear \citep{zeng2023transformers}, LightGBM \citep{ke2017lightgbm}, Chronos \citep{ansari2024chronos}, and iTransformer \citep{liu2023itransformer}, all of which generate forecasts solely from past time-series observations. These models cover a range of forecasting backbones, from classical machine learning methods to recent deep neural architectures, thereby providing a comprehensive benchmark for evaluating the robustness and general applicability of our method. The second group includes a direct prediction method built upon LoRA \citep{hu2022lora}. Unlike the history-only baselines, this model takes as input not only the historical sequence but also the same external textual information used in our approach, enabling a more direct comparison with our method. Including this setting allows for a fairer and more informative comparison, since both methods have access to the same sources of information and differ primarily in how that information is utilized for forecasting.

\subsubsection{Reward Model}
Our framework employs two complementary reward models to guide the news-augmented forecasting pipeline: an importance-aware reward model for evaluating individual news segment utility, and a Process Reward Model (PRM) [Lightman et al., 2023a] for identifying optimal news selection strategies.

\textbf{Importance-aware reward model.}
The reward model is built upon Qwen3-8B ~\citep{lightman2023letsverifystepstep} and fine-tuned via Low-Rank Adaptation (LoRA) ~\citep{hu2021loralowrankadaptationlarge} with rank $r{=}64$, scaling factor $\alpha{=}128$, and dropout rate $0.05$. LoRA adapters are applied to all attention projections ($W_Q, W_K, W_V, W_O$) as well as the feed-forward sub-layers ($W_{\text{gate}}, W_{\text{up}}, W_{\text{down}}$). A linear regression head maps the final hidden state to a scalar importance score $\rho \in [0,1]$. Training uses supervised regression with mean squared error:
\begin{equation}
    \mathcal{L}_{\text{RM}} = \frac{1}{n}\sum_{i=1}^{n}\big(R_\theta(N_i) - \rho_i\big)^2,
\end{equation}
where $\rho_i$ is the normalized importance label for article $N_i$. To obtain supervision, we compare forecasting performance under different retention rates for the same article. Let $\text{RMSE}_{\text{high}}$ denote the validation RMSE when article $N_i$ is aggressively compressed and $\text{RMSE}_{\text{low}}$ the RMSE when it is lightly compressed. We define the unnormalized utility as $\tilde{\rho}(N_i) = (\text{RMSE}_{\text{high}} - \text{RMSE}_{\text{low}})/\text{RMSE}_{\text{no\,news}}$ and normalize it to $[0,1]$ within each training split. A larger score indicates that preserving information from the article is more beneficial for forecasting. At inference time, the reward model scores individual news segments, thereby determining the retention ratio for each segment through a monotone mapping function.


\textbf{Process Reward Model (PRM).}
While the RM evaluates single news segments in isolation, the PRM assesses the quality of composite news selection strategies. We enumerate all $2^N$ subsets of the $N$ available news segments for each sample and query the forecasting model with each subset to obtain its prediction RMSE. The reward for each subset is defined as $r = -\text{RMSE}$, so that lower forecasting error corresponds to higher reward. This exhaustive evaluation captures interaction effects among news segments that pairwise comparisons cannot reveal. The PRM is trained on these step-wise labels using the TRL framework~\citep{vonwerra2020trl}, where each step corresponds to the incremental inclusion of a news segment into the selected subset. At inference time, the PRM guides a search over candidate news combinations to identify the subset that maximizes the expected forecasting accuracy.

\begin{table}[t]
\centering
\small
\caption{Ablation results for removing external news and the reward model.}
\label{tab:ablation}
\begin{tabular}{lcccc}
\toprule
Domain & Electricity & Bitcoin  & Traffic & Exchange\\
\midrule
w/o News & 1171.16 & 1107.52 & 36.88 & 28.70 \\
w/o Reward Model & 697.13 & 2008.36 & 40.41 & 7.35 \\
Our Method & 381.58 & 886.58 & 34.66 & 5.86 \\

\bottomrule
\end{tabular}
\vspace{-15pt}
\end{table}
\subsection{Main Results}
Table~\ref{tab:results1} presents the RMSE comparison between our method and baseline models across four domains. We adopt RMSE as the primary evaluation metric because it penalizes larger prediction errors more heavily than other metrics such as MAE, making it particularly suitable for capturing the impact of sudden shifts and anomalies in time series that are often triggered by external news events. As shown in Table~\ref{tab:results1}, our method achieves the best performance across all four domains, with RMSE values of 372.32, 886.58, 34.66, and 5.86 on Electricity, Bitcoin, Traffic, and Exchange, respectively. Notably, on the Bitcoin domain, our method reduces the RMSE by 23.3\% compared to the second-best baseline (LightGBM, 1156.65), demonstrating a substantial improvement. Similarly, on the Exchange domain, our method outperforms the strongest baseline (LightGBM, 7.09) by 17.3\%. These significant performance gains in news-sensitive financial domains validate the effectiveness of incorporating external news information into time series forecasting. However, the improvements on the Electricity and Traffic domains are relatively marginal. One possible reason lies in the inherent characteristics of these two domains. Both electricity consumption and traffic flow exhibit strong periodicity and regularity governed primarily by human activity patterns (e.g., daily routines, weekly cycles, and seasonal trends), making their dynamics largely predictable from historical data alone. Consequently, these domains are less sensitive to external news events, limiting the additional predictive value that news-augmented methods can offer.

\subsection{Ablation Experiments}

To further validate the effectiveness of each component in our framework, we design two groups of ablation experiments. Specifically, we consider the following variants: (1) \textbf{w/o News}, which removes the external news input and relies solely on the large language model to perform predictions based on historical time series data; and (2) \textbf{w/o Reward Model}, which removes the reward model used to optimize the news compression ratio, allowing the large language model to compress news content in an unconstrained manner before making predictions.

Table ~\ref{tab:ablation} presents the RMSE comparison. Both variants exhibit significant performance degradation compared to our method across all domains. The inferior performance of the ``w/o News'' variant confirms that external news provides critical supplementary information beyond what is captured by historical patterns alone, particularly for news-sensitive domains where exogenous factors drive market dynamics. The degraded performance of the ``w/o Reward Model'' variant demonstrates that unguided news compression introduces noise and may discard key predictive signals, highlighting the necessity of our reward model in learning an optimal compression strategy that preserves task-relevant information while filtering out irrelevant content. These results collectively validate that both the news integration mechanism and the reward-model-guided compression are indispensable components of our framework, each contributing meaningfully to the final forecasting performance.

\subsection{Convergence Analysis}

To demonstrate that our PRM can accelerate news selection, we conduct a convergence analysis. We set a predefined RMSE threshold for each domain and compare two strategies: (1) PRM-guided selection refers to selecting the news article most likely to yield the largest reduction in RMSE, where the PRM is a previously trained model that can recognize similar patterns. (2) Naive selection, which adds news without guidance. At each step, one article is appended, compressed, and used for prediction. We record the steps required to reach the target threshold.

As shown in Table~\ref{tab:convergence}, the PRM-guided strategy consistently requires fewer steps across all four domains. On average, PRM-guided selection reduces the number of convergence steps by 24.8\% compared to naive selection. The most pronounced improvement is observed in the Bitcoin domain, where the required steps decrease by 37.6\%. This confirms that our PRM effectively identifies the news most likely to improve forecasting accuracy, reducing retrieval and inference iterations. In practice, this yields lower computational cost and faster convergence, making our framework more efficient and scalable for large candidate news pools.

\begin{table}[t]
\centering
\small
\caption{Convergence comparison (number of steps to reach target RMSE) between PRM-guided and naive news selection across four domains.}
\label{tab:convergence}
\begin{tabular}{lccccc}
\toprule
Domain & Electricity & Bitcoin  & Traffic & Exchange & Avg.\\
\midrule
Naive selection & 2.44 & 2.58 & 1.40 & 2.45 & 2.22 \\
PRM-guided selection & 2.13 & 1.61 & 1.36 & 1.59 & 1.67 \\
\midrule
Improvement (\%) & 12.7\% & 37.6\% & 2.9\% & 35.1\% & 24.8\% \\
\bottomrule
\end{tabular}
\vspace{-15pt}
\end{table}

\section{Conclusion and Future Work}
In this paper, we proposed a news-augmented time series forecasting framework that addresses two practical challenges in existing LLM-based pipelines: the difficulty of incorporating long news under limited context budgets and the lack of quality control in iterative supplementary news retrieval. To this end, we introduced an importance-aware fusion module to allocate compression budgets according to the forecasting utility of each article, and a PRM-guided refinement mechanism to rank supplementary news candidates based on their expected contribution to error reduction. Experimental results on electricity, bitcoin, traffic, and exchange benchmarks demonstrate that our method consistently improves forecasting accuracy over strong baselines, while also reducing the number of refinement steps required for convergence.

For future work, we plan to extend the framework in several directions. First, we will investigate more efficient and scalable fusion strategies for larger news pools and longer forecasting horizons. Second, we aim to generalize the method to multimodal exogenous information, such as social media, reports, and event graphs, beyond plain news text. We believe these directions can further enhance the practicality and generality of LLM-based forecasting systems.

\small
\bibliography{06_references}
\bibliographystyle{iclr2026_conference}
\newpage

\appendix
\section{Appendix}
\subsection{Experimental Setting / Details}
\textbf{Analysis of Long News}

Table \ref{tab:news_length} reports the volume of news text associated with each forecasting task across the four domains. On average, each task is paired with 27–38 relevant articles, and the aggregate character count ranges from approximately 100k to 144k characters (roughly 101k–144k tokens). In the worst case, a single task can be associated with up to 601k characters for Electricity and 571k characters for Traffic. These figures far exceed the effective context windows of most contemporary LLMs (e.g., 8k–128k tokens), empirically confirming the motivation stated: naïvely concatenating all retrieved articles into the prompt is infeasible for a substantial fraction of tasks. The problem is especially acute for Electricity and Traffic, where the maximum total length approaches or exceeds 500k tokens. This underscores the necessity of our importance-aware compression module, which selectively preserves forecasting-relevant content while respecting a fixed context budget. It also highlights why the process reward model (PRM) for retrieval supervision is beneficial: when dozens of candidate articles compete for limited context space, intelligently sequencing and selecting supplementary news—rather than retrieving them in a single unguided pass.
\begin{table}[htbp]
  \centering
  \caption{Details of News Length}
  \label{tab:news_length}
  \resizebox{\columnwidth}{!}{%
  \begin{tabular}{lcccc}
    \toprule
    & Electricity & Bitcoin & Traffic & Exchange \\
    \midrule
    Avg. \# of articles per task   & 36.4 & 36.5 & 26.9 & 37.6 \\
    Avg. total chars per task       & 129,975 ($\approx$130k) & 144,016 ($\approx$144k) & 100,553 ($\approx$101k) & 111,593 ($\approx$112k) \\
    Max total chars per task        & 601,319 ($\approx$601k) & 293,829 ($\approx$294k) & 571,076 ($\approx$571k) & 561,505 ($\approx$562k) \\
    \bottomrule
  \end{tabular}
  }
\end{table}

\textbf{Baselines.}

We benchmark our method against a broad range of established baselines spanning different forecasting paradigms—including statistical models, Transformer-based architectures, and approaches leveraging large language models (LLMs). Specifically, the baselines include N-BEATS \citep{oreshkin2019n}, PatchTST \citep{nie2022time}, DLinear \citep{zeng2023transformers}, LightGBM \citep{ke2017lightgbm}, Chronos \citep{ansari2024chronos}, iTransformer \citep{liu2023itransformer}, Time-LLM \citep{jin2023time}, and LoRA-based LLM fine-tuning \citep{wang2024news}. For Chronos \citep{ansari2024chronos}, we adopt the "amazon/chronos-bolt-mini" variant. The implementations of iTransformer \citep{liu2023itransformer}, Time-LLM \citep{jin2023time}, and DLinear \citep{zeng2023transformers} are based on the neuralforecast library (https://github.com/Nixtla/neuralforecast). All baselines are configured in accordance with the original architectural specifications and implementation protocols described in their respective publications or official codebases, and hyperparameters are tuned following the authors' recommended guidelines to ensure fair and competitive performance.

\textbf{Reward Model Fine-tuning Details}

Both the Outcome Reward Model (ORM) and the Process Reward Model (PRM) are fine-tuned from Qwen3-8B using the Bradley--Terry pairwise ranking loss $\mathcal{L} = -\log\sigma(r_{\text{chosen}} - r_{\text{rejected}})$, implemented via the TRL \texttt{RewardTrainer} framework. We apply Low-Rank Adaptation (LoRA) to reduce the number of trainable parameters while preserving the pretrained representations. The LoRA configuration is shared across both models: rank $r=64$, scaling factor $\alpha=128$, and dropout rate $0.07$. Adapters are inserted into all attention projections ($W_Q$, $W_K$, $W_V$, $W_O$) and feed-forward MLP layers (gate, up, and down projections), resulting in approximately 1.2\% trainable parameters relative to the full model.

Both models are trained in bfloat16 mixed precision with the AdamW optimizer at a peak learning rate of $1.5 \times 10^{-5}$. The ORM is trained for 1{,}068 optimizer steps on approximately 15k preference pairs constructed from pairwise news importance rankings, while the PRM is trained for 800 steps on pairs derived from offline reflection logs with a composite outcome score combining RMSE improvement and directional accuracy. The maximum input sequence length for the PRM is set to 2{,}048 tokens. For both models, we evaluate every 50 training steps and select the checkpoint with the lowest validation loss as the final model. All training is conducted on four NVIDIA RTX 5880 Ada Generation GPUs with 48GB memory each.

\textbf{Evaluation.} 

To evaluate the performance of the proposed model, we adopt a commonly used error metric: Root Mean Squared Error (RMSE). RMSE computes the square root of the average squared errors, assigning a higher penalty to large deviations between predictions and true values:
\[
\mathrm{RMSE} = \sqrt{ \frac{1}{N} \sum_{i=1}^{N} \left( y_i - \hat{y}_i \right)^2 }
\]

\subsection{Example Inputs for Fine-tuning}

We provide representative examples of the input text fed to the reward models during training. Each training sample consists of a \textit{chosen} and a \textit{rejected} text; the model learns to assign a higher scalar reward to the chosen input.
\paragraph{ORM Input Example.}

The ORM receives a news article paired with its ranking context. A training pair consists of a more important (chosen) and a less important (rejected) news item:
\begin{quote}
\small
\texttt{[Chosen]:} ``Fed signals pause in rate hikes amid cooling inflation data. The Federal Reserve held interest rates steady at 5.25\%--5.50\% and indicated that future decisions will be data-dependent, citing moderating CPI and stable employment figures.''
\vspace{0.5em}
\texttt{[Rejected]:} ``Local bakery chain opens third location downtown. The family-owned business plans to hire 12 additional staff members for the new storefront.''
\end{quote}
\paragraph{PRM Input Example.}

The PRM evaluates intermediate reflection steps during the iterative news selection process. Each input encodes the current selection state ($\tau_{<t}$) and a candidate action:
\begin{quote}
\small
\texttt{You are scoring one reflection suggestion for news selection.}
\texttt{\#\# Historical Context}\\
\texttt{The target series shows a rising trend over the past 30 days with increased volatility in the recent week. Current RMSE of baseline forecast: 0.0342.}
\texttt{\#\# Already Selected News ($\tau_{<t}$)}\\
\texttt{These news already exist in the current strategy. Evaluate candidate c only by its incremental value beyond this history. Do not reward repeated information.}\\
\texttt{1. Fed holds rates steady, signals data-dependent path forward.}\\
\texttt{2. US CPI drops to 3.1\% year-over-year, below consensus estimate.}
\texttt{\#\# Candidate Suggestion}\\
\texttt{- Subset name: macro\_employment}\\
\texttt{- Subset size: 3}\\
\texttt{- Reflection constraints:}\\
\texttt{\quad - Must provide new directional signal not covered by existing selections.}\\
\texttt{- Suggested added news:}\\
\texttt{\quad 1. Non-farm payrolls miss expectations at 150K vs 180K forecast.}\\
\texttt{\quad 2. Unemployment rate ticks up to 3.9\%.}\\
\texttt{\quad 3. Average hourly earnings growth slows to 0.2\% month-over-month.}
\texttt{\#\# Agent Forecast Under This Suggestion}\\
\texttt{Predicted values: 1.042, 1.038, 1.035, 1.031, ...}
\texttt{Task: Score candidate c by incremental value over $\tau_{<t}$ (higher score = better additional information and less redundancy).}
\end{quote}

\section{Full Prompt Design (Take Bitcoin as an example)}

\subsection{ORM Data Generation}

\begin{tcolorbox}[colback=gray!5, colframe=gray!50, title=Prompt Template, fonttitle=\bfseries, breakable]
\small
\begin{verbatim}
<Historical Price Data>
----- Task Description -----
You are a financial forecasting assistant specialized
in cryptocurrency markets.
Your task is to directly predict the next 15-period
Bitcoin prices.
Based on the given historical data and any provided
context, forecast the price for each of the next
15 periods.
Notation rules:
1. Predict 15 real-valued Bitcoin prices for the
   next 15 periods.
2. Output ONLY the predicted numeric values.
3. Output must be a single line string.
4. The 15 numbers must be separated by commas.
5. Do NOT provide any explanation, reasoning, labels,
   or additional text.
6. Do NOT add spaces before or after commas.
Output format example:
58234.12,58401.55,58610.23,58790.11,58920.44,
59005.67,59120.89,59340.12,59450.33,59600.78,
59742.15,59880.50,60010.22,60145.88,60280.45
Additional market context:
1. <News Segment 1>
2. <News Segment 2>
...
N. <News Segment N>
Now predict the next 15 Bitcoin prices:
\end{verbatim}
\end{tcolorbox}
\noindent In the ablation study, we evaluate three configurations:
\begin{itemize}
    \item \textbf{Baseline}: Only historical price data and task description, without any news context (the ``Additional market context'' section is omitted).
    \item \textbf{All News}: All available news segments are included as additional market context.
    \item \textbf{Leave-One-Out}: Each news segment is excluded in turn while retaining all others, to assess the individual contribution of each news segment to forecasting accuracy.
\end{itemize}

\subsection{PRM Training Data Generation}
\label{appendix:prm-data}

To construct training data for the Process Reward Model (PRM), we perform exhaustive subset enumeration over the news segments associated with each data sample. Specifically, for a sample with $N$ news segments, we enumerate all $2^N$ subsets of the news set (including the empty set). For each subset $S \subseteq \{1, 2, \ldots, N\}$, we:

\begin{enumerate}
    \item Construct a prompt by including only the news segments indexed by $S$ in the ``Additional market context'' section of the prompt template.
    \item Query the LLM to obtain 15-step price predictions.
    \item Compute the RMSE between the predicted and ground-truth prices.
    \item Assign a reward signal $r = -\text{RMSE}$ to the subset.
\end{enumerate}

\noindent This procedure yields a labeled dataset of $(\text{subset}, \text{reward})$ pairs for each sample, which is used to train the PRM to evaluate the quality of different news subset selections. Each record in the training data contains the subset composition, the corresponding prompt, the model's predictions, and the computed reward.

\subsection{Prediction Prompt}
\label{appendix:prediction-prompt}
The following prompt is used to query DeepSeek-V3.2 for Bitcoin price forecasting. The prompt consists of historical price data, statistical summaries, task instructions, consistency constraints, and compressed news context.
\begin{tcolorbox}[colback=gray!5, colframe=gray!50, title=Prediction Prompt Template, breakable]
\small
\texttt{\{historical\_data\}}\\
\textbf{Historical data statistics:}\\
- Max: \texttt{\{hist\_max\}}\\
- Min: \texttt{\{hist\_min\}}\\
- Mean: \texttt{\{hist\_mean\}}\\
- Std: \texttt{\{hist\_std\}}\\
- Range: \texttt{\{hist\_range\}}\\
\textbf{----- Task Description -----}\\
You are a forecasting assistant specialized in Bitcoin price prediction. Your task is to directly predict the next 15 daily Bitcoin price values (in USD). Based on the given historical data and any provided news context, forecast the price for each of the next 15 days.\\
\textbf{Notation rules:}
\begin{enumerate}
    \item Predict 15 real-valued price values for the next 15 days.
    \item Output ONLY the predicted numeric values.
    \item Output must be a single line string.
    \item The 15 numbers must be separated by commas.
    \item Do NOT provide any explanation, reasoning, labels, or additional text.
    \item Do NOT add spaces before or after commas.
\end{enumerate}
\textbf{Output format example:}\\
\texttt{11362.0,11192.3,11245.4,11580.2,12858.9,11594.9,10864.8,\\10903.2,11452.3,11203.0,11118.1,11467.5,11795.1,11244.8,10166.0}\\
\textbf{IMPORTANT:} Your predictions should remain consistent with the historical data pattern. Unless the news explicitly indicates a major disruptive event (e.g., major regulation changes, exchange hacks, institutional adoption/ban, macroeconomic crisis), your predictions should stay within or close to the historical range [\texttt{\{hist\_min\}}, \texttt{\{hist\_max\}}] with mean around \texttt{\{hist\_mean\}}. Do NOT introduce large deviations without strong justification from the news.\\
\textbf{Additional context from recent news:}\\
1. \texttt{\{compressed\_news\_1\}}\\
2. \texttt{\{compressed\_news\_2\}}\\
\quad $\vdots$\\
5. \texttt{\{compressed\_news\_5\}}\\
Now predict the next 15 Bitcoin price values:
\end{tcolorbox}
\subsection{News Compression Prompt}
\label{appendix:compression-prompt}
The following prompt is used to compress news articles before feeding them into the prediction model. The target length is controlled by the allocated token budget.
\begin{tcolorbox}[colback=gray!5, colframe=gray!50, title=News Compression Prompt Template]
\small
Please compress the following news article into a concise summary that retains the most important information relevant to Bitcoin price forecasting. Focus on facts that could affect Bitcoin price (regulation, adoption, market sentiment, institutional investment, macroeconomic events, etc.). Be as concise as possible.\\
\textbf{News article:}\\
\texttt{\{news\_text\}}\\
\textbf{Compressed summary:}
\end{tcolorbox}
\subsection{Reward Model Prompt}
\label{appendix:reward-prompt}
The following prompt is used to score news relevance with the fine-tuned reward model (Qwen3-8B with LoRA adapter).
\begin{tcolorbox}[colback=gray!5, colframe=gray!50, title=Reward Model Prompt Template]
\small
\textbf{System:} You are an expert cryptocurrency analyst evaluating the relevance of news segments for predicting Bitcoin price movements.\\
\textbf{User:}\\
Historical Bitcoin price data (15 points, daily intervals, covering 15 days):\\
\texttt{\{price\_history\}}\\
News segment:\\
\texttt{\{news\_text\}}\\
How important is this news for forecasting the next 15 days of Bitcoin price? Rate it.
\end{tcolorbox}

\subsection{Process Reward Model (PRM) Prompt}
\label{appendix:prm-prompt}

The following prompt template is used by the Process Reward Model to score candidate news articles at each greedy selection step. The PRM evaluates the incremental value of each candidate given the already-selected news and the current forecast performance (RMSE).

\begin{tcolorbox}[colback=gray!5, colframe=gray!50, title=PRM Scoring Prompt Template, breakable]
\small
You are scoring one reflection suggestion for news selection.\\

\textbf{\#\# Historical Context}\\
\texttt{\{instruction\}}\\

\textbf{\#\# Already Selected News (tau<t)}\\
These news already exist in the current strategy. Evaluate candidate c only by its incremental value beyond this history. Do not reward repeated information.\\
1. \texttt{\{selected\_news\_1\}}\\
2. \texttt{\{selected\_news\_2\}}\\
\quad $\vdots$\\

\textbf{\#\# Candidate Suggestion}\\
- Subset size: 1\\
- Suggested added news:\\
\quad 1. \texttt{\{candidate\_news\}}\\

\textbf{\#\# Agent Forecast Under This Suggestion}\\
Current forecast RMSE: \texttt{\{last\_rmse\}}\\

Task: Score candidate c by incremental value over tau<t (higher score = better additional information and less redundancy).
\end{tcolorbox}

\noindent
At each greedy step $t$, the PRM scores all remaining candidate news articles $c \in \mathcal{C} \setminus \mathcal{S}_{t-1}$ given the selected history $\mathcal{S}_{t-1}$ and the RMSE from the previous prediction round. The candidate with the highest score is selected:
\begin{equation}
    c_t^* = \arg\max_{c \in \mathcal{C} \setminus \mathcal{S}_{t-1}} \text{PRM}(c \mid \mathcal{S}_{t-1},\, \text{RMSE}_{t-1})
\end{equation}
The process terminates when the normalized RMSE falls below a predefined threshold $\epsilon$ (set to 15\% in our experiments).

%
%

\graphicspath{{figures/}{analysis/figures/}}

\section{Supplementary Case Studies and Distribution Analysis}
\label{app:supplementary}


This section provides (i) a step-by-step case study contrasting PRM-guided greedy news selection with a naive LLM-based greedy baseline (Section~\ref{app:case-convergence}, Table~\ref{tab:app-exchange25-summary} and Figure~\ref{fig:app-exchange25-rmse}) and (ii) a compression case study where PRM-proportional summarization improves forecast RMSE over uniform budgeting (Section~\ref{app:case-compression}, Tables~\ref{tab:app-comp-exchange95-summary}--\ref{tab:app-comp-rba-redundant}).

\subsection{News Selection: PRM vs.\ Naive Baseline}
\label{app:case-convergence}

\begin{figure}[htbp]
  \centering
  \includegraphics[width=0.85\linewidth]{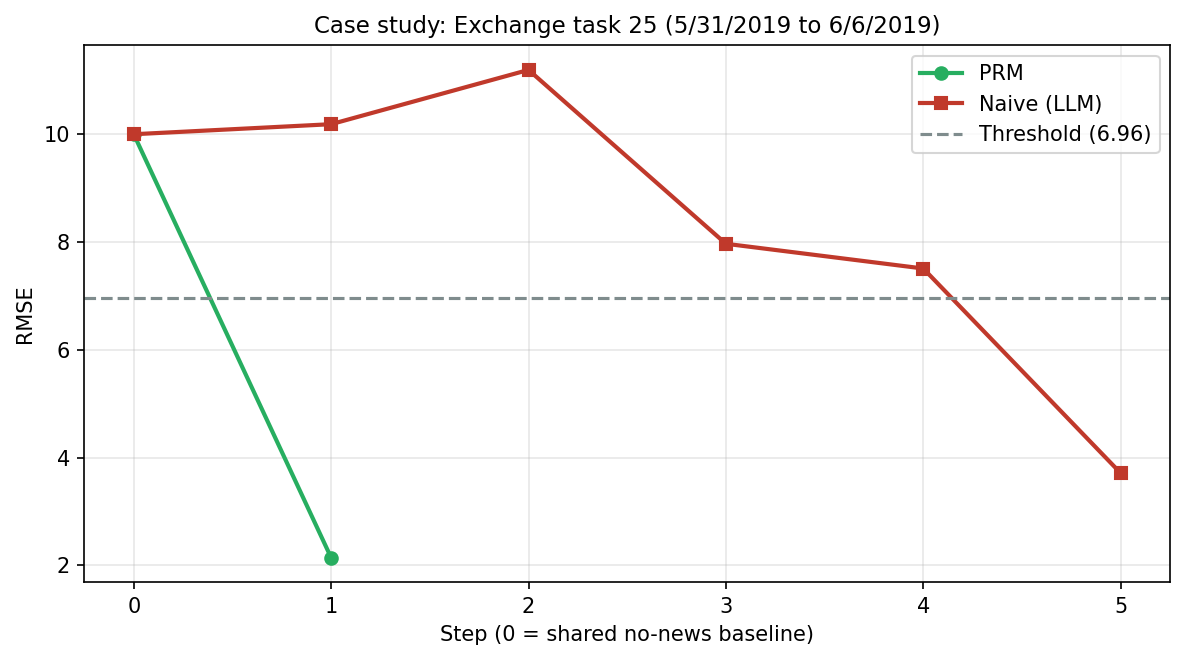}
  \caption{RMSE trace on exchange task 25. Both methods share the same step-0 (no-news) RMSE. PRM converges after one article; the naive LLM selector requires five steps.}
  \label{fig:app-exchange25-rmse}
\end{figure}

\paragraph{Setup.}
We use the exchange-rate forecasting task (AUD/USD, 7-day horizon) with task ID~25 and forecast window 5/31/2019--6/6/2019.
Both methods share the same pool of $N{=}10$ filtered news articles.
Convergence is defined as forecast RMSE falling below $\tau = 0.01 \times \bar{y} \approx 6.96$, where $\bar{y}$ is the mean ground-truth exchange rate ($696.28$).
The \textbf{naive baseline} uses the \emph{same} forecaster LLM (DeepSeek-V3.2) to greedily pick one article per step from the remaining pool, given historical data, already selected news, and the current RMSE.
\textbf{PRM greedy} instead scores all remaining candidates with the trained PRM (conditioned on selected history and \texttt{last\_rmse}), picks the highest-scoring article, and performs one forecast call.
At \textbf{step~0}, both methods share the \emph{same} no-news forecast and thus identical baseline RMSE (Table~\ref{tab:app-exchange25-summary}, Figure~\ref{fig:app-exchange25-rmse}).

\paragraph{Summary.}
The failure of the naive LLM selector occurs at \textbf{Step~1}: it chooses a low-utility geopolitical headline (PRM rank 9/10) before domestic macro/housing signals, requiring five articles to converge, whereas PRM selects the top-ranked article first and converges in one step.
Table~\ref{tab:app-exchange25-summary} summarizes the gap; Figure~\ref{fig:app-exchange25-rmse} plots the RMSE traces.

\begin{table}[htbp]
  \centering
  \caption{Exchange task 25: convergence comparison (PRM vs.\ naive).}
  \label{tab:app-exchange25-summary}
  \small
  \begin{tabular}{@{}lccp{5.2cm}@{}}
    \toprule
    Method & Steps to converge & RMSE after 1st article & First article selected \\
    \midrule
    PRM greedy & 1 & 2.14 (\checkmark\ below $\tau$) &
      \#0: Melbourne climate protest (die-in) \\
    Naive (LLM greedy) & 5 & 10.19 (above $\tau$) &
      \#8: China rare-earth supply threat (PRM rank 9/10) \\
    \bottomrule
  \end{tabular}
\end{table}

\paragraph{Step 0 (no news).}
As shown in Figure~\ref{fig:app-exchange25-rmse}, both methods start from the \textbf{same} no-news forecast with RMSE $10.0$ at step~0.

\paragraph{Step 1 (root cause).}
Table~\ref{tab:app-prm-ranking} ranks all ten candidates by PRM score at the first selection step.
\textbf{Naive LLM} selects index~8 first:
\begin{quote}
  \small
  \textit{China threatens to strangle supply of important rare earth minerals to the US.}
\end{quote}
\begin{itemize}[nosep,leftmargin=*]
  \item RMSE after step~1: $10.19$ (threshold $6.96$) --- not converged.
  \item PRM rank: \textbf{9th of 10} in Table~\ref{tab:app-prm-ranking} (score $1.0234$).
  \item Interpretation: salient geopolitical content with weak short-horizon linkage to AUD/USD vs.\ domestic RBA, housing, and protest shocks preferred by PRM.
\end{itemize}

\textbf{PRM} adds index~0 first (highest score $1.2969$):
\begin{quote}
  \small
  \textit{Thousands of climate protesters attend Melbourne `die-in'.}
\end{quote}
\begin{itemize}[nosep,leftmargin=*]
  \item RMSE after step~1: $2.14$ --- \textbf{converged} (Figure~\ref{fig:app-exchange25-rmse}).
\end{itemize}

\begin{table}[htbp]
  \centering
  \caption{PRM scores for all candidates at step~1 (exchange task 25).}
  \label{tab:app-prm-ranking}
  \footnotesize
  \begin{tabular}{@{}clrp{6.8cm}@{}}
    \toprule
    Rank & Idx & Score & Title (abbrev.) \\
    \midrule
    1  & 0 & 1.2969 & Thousands of climate protesters attend Melbourne `die-in' \\
    2  & 1 & 1.2109 & Donald Trump trade war with China could impact Aussie housing \\
    3  & 6 & 1.1953 & Deadline set for Adani's contentious coal mine \\
    4  & 3 & 1.1875 & Latest public housing property sale shows buyer confidence \\
    5  & 5 & 1.1875 & First homebuyer interest spikes as rate cut looms \\
    6  & 2 & 1.1484 & First homebuyer interest spikes (duplicate headline) \\
    7  & 9 & 1.1016 & Homebuyers' appetites return post-federal election \\
    8  & 4 & 1.0547 & House prices fast recovery after election (CoreLogic) \\
    9  & 8 & 1.0234 & China threatens rare-earth supply to US \textbf{(naive 1st pick)} \\
    10 & 7 & 1.0156 & RBA could cut cash rate to 0.5\%, JP Morgan says \\
    \bottomrule
  \end{tabular}
\end{table}

\begin{table}[htbp]
  \centering
  \caption{Naive LLM-greedy path: full RMSE trace until convergence (exchange task 25).}
  \label{tab:app-naive-trace}
  \small
  \begin{tabular}{@{}clrp{2.2cm}@{}}
    \toprule
    Step & News idx & Title (abbrev.) & RMSE after \\
    \midrule
    1 & 8 & China rare-earth supply threat & 10.19 \\
    2 & 1 & Trump trade war vs.\ Aussie housing & 11.19 \\
    3 & 5 & First homebuyer interest / rate cut & 7.97 \\
    4 & 6 & Adani coal mine deadline & 7.51 \\
    5 & 9 & Homebuyers' appetites post-election & 3.71 \checkmark \\
    \bottomrule
  \end{tabular}
\end{table}

Table~\ref{tab:app-naive-trace} lists the full naive path until convergence at step~5.

\paragraph{Takeaways.}
\begin{enumerate}[nosep,leftmargin=*]
  \item Both methods see the \emph{same} candidate pool; the gap is \textbf{selection order}, not pool construction (Tables~\ref{tab:app-exchange25-summary}--\ref{tab:app-naive-trace}).
  \item Without PRM, LLM-only greedy selection can front-load articles ranked 9/10 in Table~\ref{tab:app-prm-ranking}, delaying convergence from 1 to 5 steps (Figure~\ref{fig:app-exchange25-rmse}).
  \item PRM incremental scoring with \texttt{last\_rmse} feedback aligns news choice with forecast utility rather than headline salience alone.
\end{enumerate}

\subsection{Long-Text Compression: Uniform vs.\ Reward-Proportional Summaries}
\label{app:case-compression}

\paragraph{Setup (exchange-rate forecasting, task ID~95, 6/8--6/14/2019).}
Five news articles are selected per task (AUD/USD, 7-day horizon) during the RBA's June 2019 rate-cut week.
\textbf{Naive compression} assigns each article the same relative token budget (20\% of original length, capped at 400 tokens).
\textbf{PRM-proportional compression} allocates a fixed 1000-token budget across articles in proportion to reward-model scores (minimum 50, maximum 400 tokens per article).
Both use the same summarization prompt; only the per-article token cap differs.

\paragraph{Forecast outcome.}
PRM-proportional compression reduces forecast RMSE by roughly half on this task (Table~\ref{tab:app-comp-exchange95-summary}): $8.89$ (naive) vs.\ $4.45$ (PRM).
Both methods produce plausible 7-day levels near $\sim$690--698; the gain comes from rebalancing tokens away from a redundant long post-cut headline toward higher-reward pre-cut policy signals (Examples~A--B).

\begin{table}[htbp]
  \centering
  \caption{Exchange task 95: forecast RMSE under naive vs.\ PRM-proportional compression.}
  \label{tab:app-comp-exchange95-summary}
  \small
  \begin{tabular}{@{}lrr@{}}
    \toprule
    Method & RMSE & Relative change \\
    \midrule
    Naive (uniform 20\%) & 8.89 & --- \\
    PRM-proportional & 4.45 & $-50\%$ \\
    \bottomrule
  \end{tabular}
\end{table}

\subsubsection{High-reward article (pre-cut policy signal)}
\label{app:comp-treasurer}

Table~\ref{tab:app-comp-treasurer} contrasts budgets for the highest-scoring headline; summaries follow.

\begin{table}[htbp]
  \centering
  \caption{Compression budgets: Treasurer urges banks to pass through rate cuts (exchange task 95).}
  \label{tab:app-comp-treasurer}
  \small
  \begin{tabular}{@{}lrrr@{}}
    \toprule
    & Reward score & Naive budget (tokens) & PRM-proportional (tokens) \\
    \midrule
    Value & $+3.11$ & 135 (20\%) & 270 (40\%) \\
    \bottomrule
  \end{tabular}
\end{table}

\noindent\textbf{Naive uniform summary:}
\begin{quote}
  \footnotesize
  The RBA is widely expected to cut its cash rate from 1.5\% to 1.25\% on Tuesday to stimulate a weak economy. Supporting this move, recent data shows soft growth, with Q1 GDP forecast at just 0.4\%. This dovish monetary policy stance and weak economic fundamentals are negative for the AUD.
\end{quote}

\noindent\textbf{PRM-proportional summary:}
\begin{quote}
  \footnotesize
  RBA expected to cut cash rate to 1.25\% to stimulate weak economy. Weak Q1 GDP growth forecast (0.4\%) and low inflation justify the dovish move. This monetary easing is likely to put downward pressure on the AUD.
\end{quote}

\noindent\textbf{Why this matters:}
PRM \emph{doubles} the token budget on the top-ranked article (Table~\ref{tab:app-comp-treasurer}), emphasizing the \emph{pre-cut} policy stance (Treasurer pressuring banks) rather than treating it like any other 20\% slice.

\subsubsection{Low-reward article (redundant long RBA headline)}
\label{app:comp-rba-redundant}

The post-announcement ``RBA slashes rate'' piece is long ($\sim$2535 tokens) but scores near zero for incremental forecast utility; Table~\ref{tab:app-comp-rba-redundant} shows the budget shift.

\begin{table}[htbp]
  \centering
  \caption{Compression budgets: ``RBA slashes cash rate to all-time low'' (exchange task 95).}
  \label{tab:app-comp-rba-redundant}
  \small
  \begin{tabular}{@{}lrrr@{}}
    \toprule
    & Reward score & Naive budget (tokens) & PRM-proportional (tokens) \\
    \midrule
    Value & $+0.15$ & 400 (20\%, max cap) & 202 (8\%) \\
    \bottomrule
  \end{tabular}
\end{table}

\noindent\textbf{Naive uniform summary:}
\begin{quote}
  \footnotesize
  RBA cuts cash rate 25bps to record low 1.25\% to support employment and inflation. AUD/USD rose slightly to ${\sim}0.6976$, as markets had fully priced in the cut and some expected a larger move. Economists forecast further rate cuts ahead, with weak domestic data (low inflation, rising unemployment) and global trade tensions cited as key reasons. This dovish outlook is typically negative for the AUD.
\end{quote}

\noindent\textbf{PRM-proportional summary:}
\begin{quote}
  \footnotesize
  RBA cuts cash rate 25bps to record low 1.25\% to support employment and inflation. Further cuts are anticipated, with markets expecting a total reduction to 0.5\% by 2020. Despite the cut, AUD/USD rose slightly to 0.6976, indicating the move was fully priced in and some had expected a larger cut.
\end{quote}

\noindent\textbf{Budget asymmetry:}
naive spends the \textbf{400-token maximum} on this repetitive post-cut story; PRM allocates only 202 tokens (8\% of original length), freeing budget for sharper pre-cut macro cues in Example~A.

\paragraph{Takeaways.}
\begin{enumerate}[nosep,leftmargin=*]
  \item Naive uniform compression over-funds long, low-incremental-utility headlines at the same 20\% rule as core macro drivers (Table~\ref{tab:app-comp-rba-redundant}).
  \item PRM-proportional budgeting concentrates tokens on high-reward news (Table~\ref{tab:app-comp-treasurer}) and trims redundant rate-cut coverage.
  \item On exchange task~95, this rebalancing lowers RMSE from $8.89$ to $4.45$---a clear improvement that illustrates PRM-guided compression without extreme forecast failure under the naive baseline.
\end{enumerate}

\end{document}